\documentclass{article}
\usepackage{spconf,amsmath,graphicx}
\usepackage{multirow}
\usepackage{booktabs}
\usepackage{makecell}
\usepackage{amssymb}
\usepackage{bbding}
\usepackage{color}
\usepackage{algorithm}  
\usepackage{algpseudocode}  
\usepackage{amsmath}  
\usepackage{amsfonts}
\usepackage{bbding}
\usepackage{pifont} 
\usepackage{fontawesome}

\title{Ontology-Aware Network for Zero-Shot Sketch-based Image Retrieval}
\name{Haoxiang Zhang, He Jiang, Ziqiang Wang, Deqiang Cheng* \thanks{*Coressponding author: Deqiang Cheng. This work was supported in part by the National Natural Science Foundation of China under Grant (No.52204177) and supported in part by the Fundamental Research Funds for the Central Universities (2020QN49).}}
\address{School of Information and Control Engineering, China University of Mining and Technology}
\begin{document}
\maketitle
\begin{abstract}
Zero-Shot Sketch-Based Image Retrieval (ZSSBIR) is an emerging task. The pioneering work focused on the modal gap but ignored inter-class information. Although recent work has begun to consider the triplet-based or contrast-based loss to mine inter-class information, positive and negative samples need to be carefully selected, or the model is prone to lose modality-specific information. To respond to these issues, an Ontology-Aware Network (OAN) is proposed. Specifically, the smooth inter-class independence learning mechanism is put forward to maintain inter-class peculiarity. Meanwhile, distillation-based consistency preservation is utilized to keep modality-specific information. Extensive experiments have demonstrated the superior performance of our algorithm on two challenging  Sketchy and Tu-Berlin datasets.
\end{abstract}
\section{Introduction}
ZSSBIR [1] has been popular recently, which is more challenging than Sketch-Based Image Retrieval (SBIR) due to lacking the knowledge of unseen test categories. It is well known that the modal gap between sketches and images makes it difficult for SBIR to obtain good results. However, the ZSSBIR has to consider not only the inevitable modal gap between sketches and images but also the transfer of knowledge from seen classes to unseen classes, which motivates this community to gain more attention.

Coming to the zero-shot learning-based methods, auxiliary semantic information needs to be considered to assist the model to obtain good results. Seen attribute vectors \cite{12} are projected onto a semantic similarity embedding space, where the unseen class is regarded as a mixture of seen classes. Semantic autoencoder \cite{14} is proposed with additional reconstruction constraint, proving to be a good explanation for unseen classes. When talking about the ZSSBIR family, both the knowledge transfer and domain gap should be considered. The pioneering works, such as ZSIH and PCYC \cite{2,3}, attempted a two-CNN network to preserve the association between sketch modality and real image modality, where the teacher network is employed to spread knowledge \cite{8}. Semantic preserving network \cite{4} is put forward by considering both semantic and visual features. A latent space is sought in which a cross-aligned latent learning method \cite{16} is applied to fuse multimodal features. Besides, a dual learning framework \cite{17} is utilized cyclically to map the sketch and image features to a common semantic space. However, the important thing is 
the inter-class peculiarity is not mined by the above-mention approaches. A discriminative model \cite{5} is realized by using triplet loss to bridge the domain gap through a gradient reversal layer. Data augmentation and a memory bank \cite{6} are used to eliminate intra-class variability, and a pre-trained CNN model is employed to keep knowledge. This kind of model could preserve category-level properties at the expense of modality-specific information, resulting in model suppression. Additionally, the sample selection is crucial, and the large batch size leads to expensive training costs. 

In this paper, an Ontology-Aware Network is proposed, and our contributions can be summarized as follows:\\
I. For ZSSBIR, an model is shown in Fig. 1, namely OAN, which is free of sample selection with moderate training cost.\\
II.  To keep inter-class peculiarity, the smooth inter-class independence learning mechanism is put forward. \\
III. Meanwhile, the proposed distillation-based consistency preservation can protect modality-specific information.

\begin{figure*}[h]	
	\centering
	\includegraphics[scale=0.155]{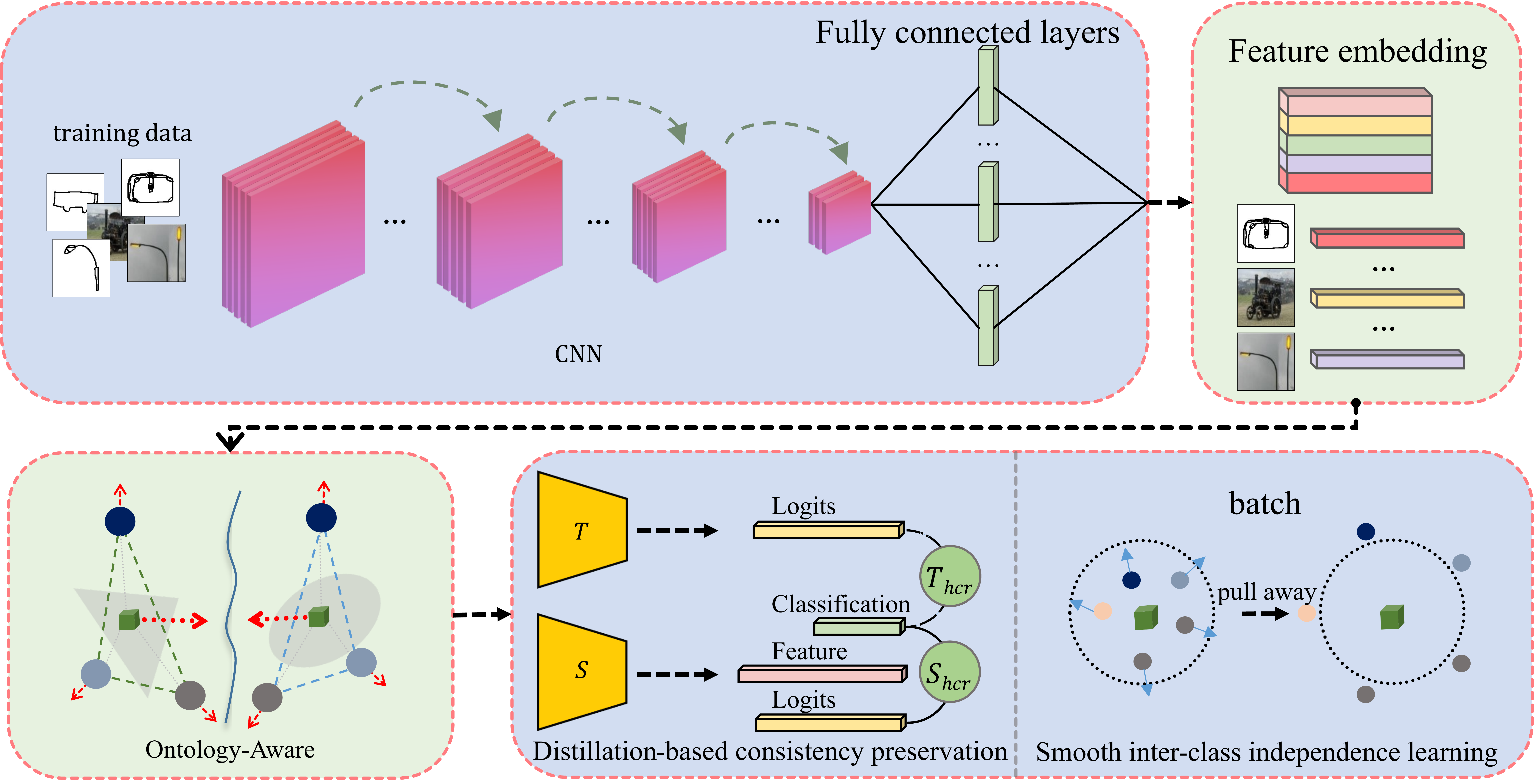}
	\caption{The general framework of our work. Firstly, CSE-ResNet50 is adopted as a feature extractor to map sketches and images to common space. Secondly, feature embedding is realized by full connection layers. Thirdly, our OAN, i.e. Ontology-Aware Network, is used to perform the smooth inter-class independence learning mechanism and distillation-based consistency preservation on embedded features, which can keep their inter-class peculiarity and modality-specific information, respectively.}
\end{figure*}

\section{Ontology-Aware Network}
The dataset in the OAN is defined as $ \mathcal{R}: \{\mathcal{R}^s,\mathcal{R}^u\} $, and the $ \mathcal{R}^s $ and $ \mathcal{R}^u $ represent the seen and unseen dataset, respectively. $ x_i^s $ and $ y_i^s $ represent the sketches and real images, where $ i\in \{s,u\} $. During the learning period, $ x^s $ and $ y^s $ are used to train the model, and $ \{x^s ,y^s\}\in \mathcal{R}^s $. Meanwhile, the unseen dataset $ \mathcal{R}^u =\{x^u ,y^u\} $ is used for testing.  

\subsection{Smooth Inter-class Independence Learning}
Recent works learned a common space for sketches and real images to cope with the modal gap and utilized triplet-based loss or contrast-based loss to maintain intra-class consistency. However, the positive and negative samples need to be carefully selected, which leads to complex mining methods. Besides, the expansion of batch size gives rise to extremely high training costs. 

Our work is distinguished from other models. In this paper, we only focus on the feature vectors within the mini-batch and treat each sketch or image ontology of the mini-batch as a core, implicitly pushing away irrelevant cores from the current ontology core. Inspired by the work [9], the smooth inter-class independence supervised learning method is proposed. Concretely, feature dictionary $ \mathbb{Q} = \{ \textbf{K}_1:\textbf{V}_1, \textbf{K}_2:\textbf{V}_2,\cdots,\textbf{K}_n:\textbf{V}_n\}  $ is created, where the $ \textbf{K} $ store the ontology weights and the $ \textbf{V} $ memorize feature vectors. In the dictionary $ \mathbb{Q} $, the key $ \textbf{K}_i $ in a mini-batch corresponds to the value $ \textbf{V}_i $. Every instance $ z_i $ including sketches $ x_i^s $ and images $ y_i^s $ is captured in the batch dataflow to create the instance feature vector set $ \textbf{V} = \{\mathcal{F}_{z_1}^s,\cdots, \mathcal{F}_{z_i}^s, \cdots \mathcal{F}_{z_n}^s| z_i\in(x_i^s,y_i^s)\}  $ for the current batch. Let $ \mathcal{F}_{z_i}^s = f(z_i), i\in \{1,2,\cdots,n\} $, where the function $ f(\cdot) $ is responsible for capturing the feature representation of the current instance. In our work, the feature vector $ \textbf{V}_i $ of instance $ i $ is mapped to 2048d. Then, $ \textbf{K}_i \gets w\textbf{K}_i + (1-w)\textbf{V}_i $ and $ \textbf{K}_i \gets \frac{\textbf{K}_i}{||\textbf{K}_i||_2} $ are adopted to update $ \textbf{K}_i $ for each $ i $, and constant $ w $ is set to 0.01.

As stated before, our smooth inter-class independence learning treats the instance ontology as a category center. To achieve the goal, the probability of each category in a batch is computed and implicitly other targets are pushed away by maximizing the inner product of ontology instances and ontology weights. At the same time, to avoid overconfidence and improve generalization, we propose smooth inter-class independence loss $ \mathcal{L}^{in} $ , which is computed as Eq.1:

\begin{equation}
	\mathcal{L}^{in} = \xi\sum_{i=1}^n{\rm log}\frac{{\rm exp} (\beta\textbf{K}_i^\top\textbf{V}_i)}{\sum_{n=1}^{N_{bc}} {\rm exp}(\beta\textbf{K}_i^\top\textbf{V}_i)}+\eta\frac{\sum {\rm log}(p_{z_i}^{pred})}{N_{cls}}
\end{equation}
, where $ N_{cls} $ denotes the batch sample numbers, and $ N_{bc} $ is the number of batch categories. The $ p_{z_i}^{pred} $ is predicted probability of instance $ i $, $ \eta $ is a  smooth parameter that can be used to improve generalization, $ \xi = -\frac{1}{N_{cls}}-\eta $, and $ \beta $ is a temperature parameter that  balances the scale of distributions.

\subsection{Distillation-based Consistency Preservation} 
In the task of ZSSBIR, the well-known way is to extract depth features from the sketches and image candidate gallery. Appointing the established Euclidean distance or other similarity metrics performs the retrieval task. To maintain feature consistency, both sketches, as well as images, are usually used as positive and negative samples. However, this approach results in the loss of modality specificity information, which in turn affects discriminability. For this reason, we propose consistency of self-distillation and teacher-student distillation that adopt the hypersphere consistency constraint \cite{7}. The feature embedding layers are trained to preserve the modal specificity. First, the paired distance between the logit layer $ \mathcal{G} $ and the classification layer $ \mathcal{C} $ can be measured as $ d_{\mathcal{V}}(z_m,z_n) = ||\mathcal{V}_{z_m}^s-\mathcal{V}_{z_n}^s||_2^2$, where  $ \mathcal{V}\in\{\mathcal{G}, \mathcal{C}\} $ and $ \mathcal{V}_{z}^s $ can be regarded as the output of logit layer or classification layer. The $ d_{\mathcal{V}}(\cdot) $ represents Euclidean distance operator and $ z_m \neq z_n $. The similarity measure can be written as $ 	D(d_\mathcal{V}) = \frac{\rho}{\delta_\mathcal{V}\sqrt{2\pi}}{\rm exp}(-\frac{(d_\mathcal{V}-\mu_\mathcal{V})^2}{2\delta_\mathcal{V}^2}) $, in which $ \delta_\mathcal{V} $ and $ \mu_\mathcal{V} $ represent variance and mean, respectively. Furthermore, $ d_\mathcal{V}\sim(0,\frac{1}{2}) $ and $ \rho $ is defined as a constant used to force the scope of $ \mathcal{D}(d_\mathcal{V}) $ within  $ [0,1] $. To maximize the similarity between $ \mathcal{D}(d_\mathcal{V}) $ and 
$ \mathcal{D}(d_\mathcal{G}) $, Eq.2 is used.

\begin{equation}
	\mathcal{L}^{\mathcal{I}_{hcr}} =-\mathcal{D}(d_{\mathcal{C}})log\mathcal{D}(*)-(1-\mathcal{D}(d_{\mathcal{C}}))log(1-\mathcal{D}(*))
\end{equation}
,where $* = d_{\mathcal{G}_{\mathcal{\mathcal{T}}}}/ d_{\mathcal{G}_{\mathcal{\mathcal{S}}}}$. $ \mathcal{L}^{\mathcal{I}_{hcr}} $  is the loss  of self-distillation part and $ \mathcal{T}_{hcr} $ or $ \mathcal{S}_{hcr} $ denote the constraint of teacher or student model, which is shown in Fig. 1. $ \mathcal{G}_{\mathcal{T}} $ and $ \mathcal{G}_{\mathcal{S}} $ represent the output of the logit layer of the $ \mathcal{T}_{hcr} $ or $ \mathcal{S}_{hcr} $, respectively.

\subsection{Classification Loss}
To help the model learn specific information well, the cross entropy loss is used, which  is computed in Eq.3. 
\begin{equation}
	\mathcal{L}^{cls} =-\frac{1}{N_{cls}}\sum_{j=1}^{N_{cls}}{\rm log}\frac{{\rm exp}(\mathcal{C}_{z_j}^{\mathcal{S}})}{\sum_{c\in \mathcal{T}^s} {\rm exp}(\mathcal{C}_{z_{c,j}}^{\mathcal{S}})}
\end{equation}
, where $ \mathcal{L}^{cls} $ represents the probability of $ \mathcal{C}_{z_j}^{\mathcal{S}} $ that the instance $ j $ in the seen domain $ \mathcal{S} $ belongs to the category $ c $, and $ \mathcal{T}^s $ denotes the number of categories in $ \mathcal{S} $ and $ N_{cls} $ denotes the number of samples in a batch. Motivated by the paper [8], we adopt the teacher with semantic information $ \mathcal{E} $ to regulate the student, which can be computed in Eq.4. 
\begin{equation}
	\mathcal{L}^{se} =-\frac{1}{N_{se}}\sum_{t=1}^{N_{se}}\sum_{k\in\mathcal{M}}\mathcal{E}_{t,k}{\rm log}\frac{{\rm exp}(\mathcal{G}_{z_t}^{\mathcal{S}})}{\sum_{q\in\mathcal{M}}{\rm exp}(\mathcal{G}_{z_{q,t}}^{\mathcal{S}})}
\end{equation}
, where $ \mathcal{L}^{se} $ represents the probability of $ \mathcal{G}_{z_{t}}^{\mathcal{S}} $ that the instance $ t $ in the semantic information, $ \mathcal{M} $ denotes the category number of semantic labels, and $ N_{se} $ is regarded as the sample number in a batch.

\subsection{The Overall Loss Function }
\begin{equation}
	\mathcal{L} = \mathcal{L}^{cls} + \lambda_1 \mathcal{L}^{se} + \lambda_2 \mathcal{L}^{in} + \lambda_3 \mathcal{L}^{\mathcal{I}_{hcr}}
\end{equation}
The overall Loss function $ \mathcal{L} $ is computed in Eq.5, where $ \lambda_1 $, $ \lambda_2 $, and $ \lambda_3 $ are the hyperparameters, and they can balance the contributions of different parts. Here, the experience of paper [8] is followed, and $ \lambda_1 $ is set to 1. 

\section{Experiment Analysis}

\subsection{Datasets} To verify the effectiveness of our OAN, two challenging datasets, i.e. Sketchy\cite{10} and Tu-Berlin\cite{11}, are employed. Sketchy includes 125 categories, 75,471 sketches, and 73,002 natural images; There are about 250 categories in Tu-Berlin, with a total of 20,000 sketches and 204,489 natural images. We follow the previous work [8] to select 100 categories in Sketchy for training and the rest is for testing. In addition, 21 categories are selected as the test set in Sketchy-B. In Tu-Berlin, 220 categories are selected for training, and other categories are used for testing. 

\subsection{Experiment Setting} 
Our method is implemented with PyTorch on RTX 3090 GPU and pre-trained CSE-ResNet50 is used to provide semantic support or logit output. In our general parameter settings, the batch size is set to 96, and the epoch is set to15. Finally, $ \lambda_1 $, $ \lambda_2 $, and $ \lambda_3 $ are set to 1, 0.001 and 0.1 in all experiments, unless otherwise stated.

\subsection{Ablation Study}
In this subsection, different modules are validated and ablation studies are carried out on Sketchy, which is shown in Table 1. Conclusions can be drawn that our method significantly improves the baseline. Specifically, benefit from the $ \mathcal{L}^{in} $, our model, i.e. OAN, does not require forced alignment in modality and only needs to focus on mini-batch, where each category is regarded as its own center. The ontology center of other categories does not belong to it, which naturally alienates the center of other categories. Similarly, by adding $ \mathcal{L}^{\mathcal{S}_{hcr}} $, the model can learn a more powerful feature representation for each category in the training process, which illustrates self-distillation makes the model stick out.  Moreover, when the $ \mathcal{L}^{\mathcal{T}_{hcr}} $ and $ \mathcal{L}^{\mathcal{S}_{hcr}} $ are considered together, metric mAP@all is slightly improved, while metric Prec@100 is not improved but decreased, and 
the overall performance of the system suffers a small but non-negligible loss, which indicates that the highly knowledgeable teacher model has obstacles in transferring knowledge to the less able student model. To sum up, only the combination of  $ \mathcal{L}^{in} +  \mathcal{L}^{\mathcal{S}_{hcr}} $ is the optimal choice of our OAN.

\begin{table}[htbp]
	\centering
	\begin{small}
		\setlength{\tabcolsep}{2.0mm}{
			\caption{Ablation study about  $ \mathcal{L}^{in} $, $ \mathcal{L}^{\mathcal{T}_{hcr}} $ and $ \mathcal{L}^{\mathcal{S}_{hcr}} $, and the best test values of objective metrics are marked in \textbf{black bold}}
			\begin{tabular}{cccccc} \hline
				Baseline [8] & $ \mathcal{L}^{in} $ & $ \mathcal{L}^{\mathcal{T}_{hcr}} $  & $ \mathcal{L}^{\mathcal{S}_{hcr}} $ & Prec@100 & mAP@all \\\hline
				\Checkmark  &  \ding{56}  &  \ding{56}  &  \ding{56}  & 0.6920 & 0.5470 \\
				\Checkmark &   \ding{56} & \ding{56}  &  \Checkmark   &0.6941 & 0.5678\\
				\Checkmark &   \Checkmark  &  \ding{56} &  \ding{56} & 0.7170 &0.5914 \\
				\Checkmark  &   \Checkmark  &  \Checkmark  & \ding{56} & 0.7174 & 0.5946 \\
				\Checkmark  &  \Checkmark  & \ding{56}& \Checkmark &  \textbf{0.7233} & 0.5994 \\
				\Checkmark  &  \Checkmark  & \Checkmark & \Checkmark &  0.7216 & \textbf{0.6008} \\
				\hline 
		\end{tabular}}
	\end{small}	
\end{table}

\begin{table*}[htbp]
	\center
	\caption{Comparisons with the existing SOTA discriminative algorithms and the best and second best results are  marked in \textbf{black bold} and \textcolor{blue}{\textbf{blue bold}}, and subscript b indicates the binary hashing results.}
	\begin{small}
		\setlength{\tabcolsep}{4.4mm}{
			\begin{tabular}{c|cccccc} \hline
				\multirow{2}{*}{Task} &\multirow{2}{*}{Methods} & \multicolumn{2}{c}{Sketchy} & \multicolumn{1}{c}{Sketchy-B} & \multicolumn{2}{c}{Tu-Berlin}  \\\cline{3-7}
				\multirow{2}{*}{} &  \multirow{2}{*}{} & mAP@all & Prec@100 &  Prec@200 & mAP@all & Prec@100 \\\hline
				\multirow{4}{*}{ZSL} & $ \rm SSE $ (ICCV'2015) [12] & 0.108 & 0.154 & - & 0.096 & 0.133 \\
				\multirow{4}{*}{} & $ \rm ZSH_b $ (ACM MM'2016) [13] & 0.165 & 0.217 & - & 0.139 & 0.174 \\
				\multirow{4}{*}{} & $ \rm SAE $ (CVPR'2017) [14] & 0.210 & 0.302 & 0.238 & 0.161 & 0.210 \\
				\multirow{4}{*}{} & $ \rm FRWGAN $ (ECCV'2018) [15] & 0.127 & 0.169 & - & 0.110 & 0.157 \\\hline
				\multirow{13}{*}{ZSSBIR} & $ \rm CAAE $ (ECCV'2018) [1] & 0.169 & 0.284 & 0.260 & - & - \\
				\multirow{13}{*}{} & $ \rm ZSIH $ (CVPR'2018) [2] & 0.258 & 0.342 & - & 0.223 & 0.294 \\
				\multirow{13}{*}{} & $ \rm PCYC $ (CVPR'2019) [3] & 0.349 & 0.463 & - & 0.297 & 0.426 \\
				\multirow{13}{*}{} & $ \rm PCYC_b $ (CVPR'2019) [3] & 0.344 & 0.399 & - & 0.293 & 0.392 \\
				\multirow{13}{*}{} & $ \rm SAKE $ (ICCV'2019) [8] & 0.547 & 0.692 & 0.598 & 0.475 & 0.599\\
				\multirow{13}{*}{} & $ \rm SAKE_b $ (ICCV'2019) [8] & 0.364 & 0.487 & 0.477 & 0.359 & 0.481 \\
				\multirow{13}{*}{} & $ \rm LCALE $ (AAAI'2020) [16] & 0.476 & 0.583 & - &  - & - \\
				\multirow{13}{*}{} & $ \rm OCEAN $ (ICME'2020) [17] & 0.462 & 0.590 & - & 0.333 & 0.467 \\
				\multirow{13}{*}{} & $ \rm StyleGuide $ (TMM'2021) [18] & 0.376 & 0.484 & 0.400 & 0.254 & 0.355 \\
				\multirow{13}{*}{} & $ \rm DSN $ (IJCAI'2021) [5] & 0.583 & 0.704 & 0.597 & 0.481 & 0.586 \\
				\multirow{13}{*}{} & $ \rm DSN_b $ (IJCAI'2021) [5] & 0.581 & 0.700 & - & 0.484 & 0.591 \\
				\multirow{13}{*}{} & $ \rm Proposed \quad OAN $ & \textcolor{blue}{\textbf{0.599}}  & \textcolor{blue}{\textbf{0.723}} & \textcolor{blue}{\textbf{0.616}}  & \textcolor{blue}{\textbf{0.500}} & \textcolor{blue}{\textbf{0.617}} \\
				\multirow{13}{*}{} & $ \rm Proposed \quad OAN_b $ & \textbf{0.617} & \textbf{0.737} & \textbf{0.621} & \textbf{0.505} & \textbf{0.625} \\\hline
		\end{tabular}}
	\end{small}	
\end{table*}

\subsection{Experiment Analysis and Visualization.} 

In view of ZSSBIR, our OAN is compared with several SOTA algorithms, such as CAAE \cite{1}, ZSIH \cite{2}, PCYC \cite{3}, DSN \cite{5}, SAKE \cite{8}, LCALE \cite{16}, OCEAN \cite{17}, StyleGuide \cite{18}, and other algorithms \cite{12,13,14,15}. As shown in Table 2, our OAN shows strong cross-modal retrieval capability. Moreover, it can produce very competitive results whether for the real image or binary image hashing. Particularly, when comes to real value retrieval, our algorithm outperforms the SAKE by about 9.5\%  in Sketchy and 5.3\% in Tu-berlin. When the feature is encoded as a binary hash value, our model receives 0.737 in the Prec@100, improving 5.3\% and 5.8\% than $ \rm DSN_b $ in Tu-berlin. Evaluated on the challenging Sketchy-B, our model comes out in front which outperforms the suboptimal algorithm by 3.0\%.

\begin{figure}[h]	
	\centering
	\includegraphics[scale=0.42]{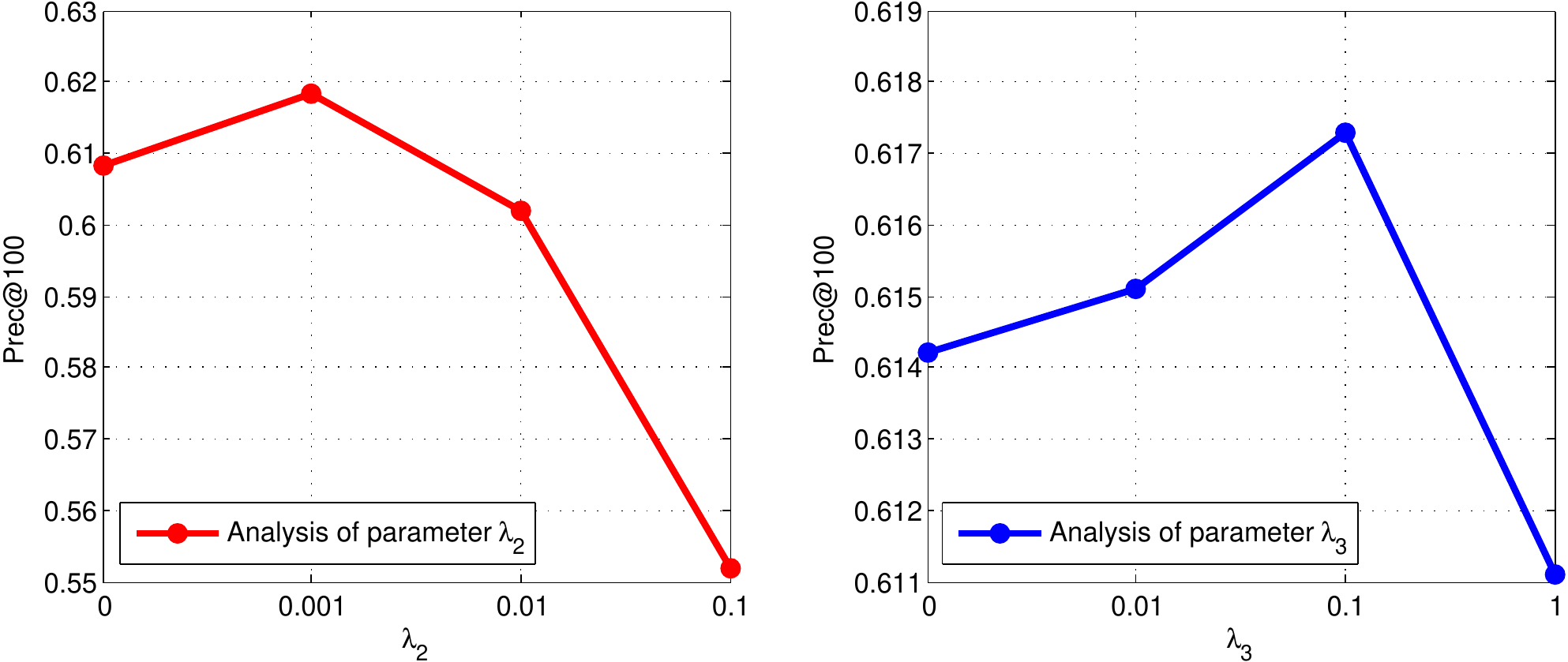}
	\caption{Analysis of parameter $ \lambda_2 $ and $ \lambda_3 $.}
\end{figure}

As is mentioned above, $ \lambda_1 $ is a constant set to 1. Therefore, the analysis of parameters is only carried out in $ \lambda_2 $ and $ \lambda_3 $. Obviously, as is seen in Fig. 2, the model achieves the best performance when the $ \lambda_2 $ and $ \lambda_3 $ are set to 0.001 and 0.1 accordingly. The top-5 retrieval results on the Tu-Berlin dataset are presented in Fig. 3 and the images in the green border are the correct retrieval results. The result of the false retrieval can be easily understood because there is a great structurally similarity between the sketch of the boat and the banana in the real image.

\begin{figure}[h]	
	\centering
	\includegraphics[scale=0.25]{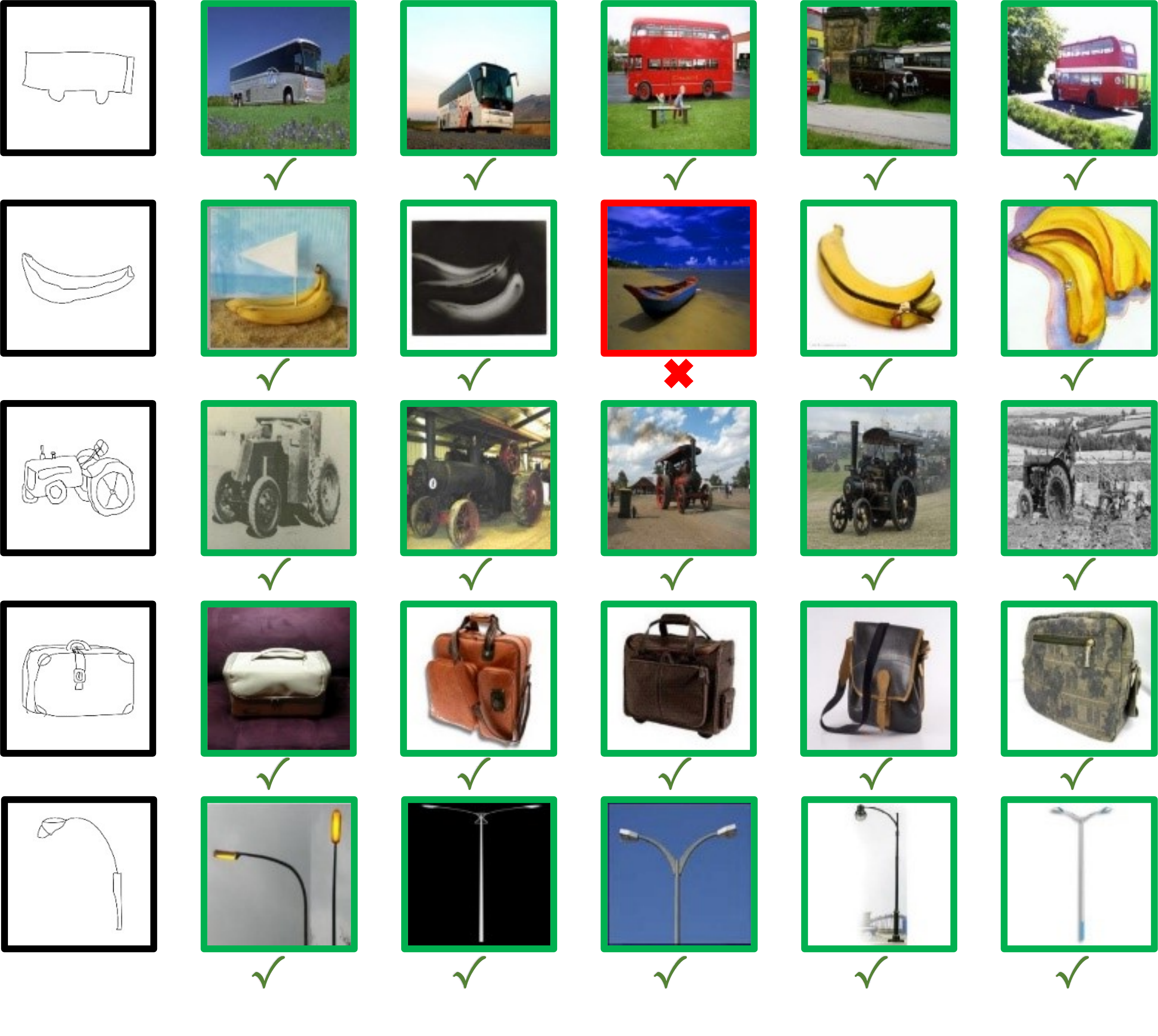}
	\caption{Visualization results.}
\end{figure}

\section{Conclusion} 
In this paper, an effective model called Ontology-Aware Network is proposed. First, the smooth inter-class independence learning mechanism is put forward to keep inter-class peculiarity. At the same time, to resist the loss of specific information, distillation-based consistency preservation is adopted for modality-specific information. Extensive experiments have proven the excellent performance of our algorithm on two challenging datasets, namely Sketchy and Tu-Berlin.

\end{document}